\documentclass[10pt,twocolumn,letterpaper]{article}
\usepackage{cvpr}
\usepackage{graphicx}
\usepackage{amsmath}
\usepackage{amssymb}
\usepackage{booktabs}
\usepackage{multirow}
\usepackage[accsupp]{axessibility}
\usepackage{calc}

\usepackage[pagebackref,breaklinks,colorlinks]{hyperref}
\usepackage[capitalize]{cleveref}
\usepackage[nolist]{acronym}
\usepackage[numbers]{natbib}
\usepackage{collect}
\usepackage{flushend}
\usepackage{adjustbox}
\usepackage{bm}
\usepackage{algorithm}
\usepackage{algpseudocode}
\usepackage{siunitx}

\usepackage[accsupp]{axessibility}

\def\cvprPaperID{12698}
\def\confName{CVPR}
\def\confYear{2024}

\usepackage{soul}
\usepackage{amsmath}
\usepackage{amssymb}
\usepackage{wrapfig}
\usepackage{booktabs}
\usepackage{multirow}
\usepackage{xspace}
\usepackage{xcolor}

\def\figurePath{images/}

\NewDocumentCommand{\rot}{O{45}
O{1em}m}{\makebox[#2][l]{\rotatebox{#1}{#3}}}%

\def\myfigure#1#2{%
    \begin{figure}[htb]
    \centering\includegraphics*[width = \linewidth]{\figurePath#1}%
    \vspace{-.2cm}%
    \caption{#2}%
    \label{fig:#1}%
    \end{figure}%
}

\def\mycfigure#1#2{%
    \begin{figure*}[htb]%
    \centering\includegraphics*[width = \linewidth]{\figurePath#1}%
    \vspace{-.2cm}%
    \caption{#2}%
    \vspace{-.2cm}%
    \label{fig:#1}%
    \end{figure*}%
}

\newcommand{\refSec}[1]{Sec.~\ref{sec:#1}}
\newcommand{\refFig}[1]{Fig.~\ref{fig:#1}}

\newcommand{\refTab}[1]{Tab.~\ref{tab:#1}}
\newcommand{\refAlg}[1]{Alg.~\ref{alg:#1}}

\soulregister\ref7
\soulregister\cite7
\soulregister\refFig7
\soulregister\refAlg7
\soulregister\cite7
\soulregister\ref7
\soulregister\pageref7
\soulregister\shortcite7
\soulregister\eg0
\soulregister\ie0
\soulregister\etal0

\DeclareGraphicsExtensions{.png,.jpg,.pdf,.ai,.psd}
\newcommand{\mysection}[2]{\section{#1}\label{sec:#2}}
\newcommand{\mysubsection}[2]{\subsection{#1}\label{sec:#2}}

\definecollection{mymaths}
\newcommand{\mymath}[2]{
    \newcommand{#1}{\TextOrMath{$#2$\xspace}{#2}}
    \begin{collect}{mymaths}{}{}{}{}
    #1
    \end{collect}
}

\definecolor{colorA}{HTML}{4285f4}
\definecolor{colorB}{HTML}{ea4335}
\definecolor{colorC}{HTML}{fbbc04}
\definecolor{colorD}{HTML}{34a853}
\definecolor{colorE}{HTML}{ff6d01}
\definecolor{colorF}{HTML}{46bdc6}
\definecolor{colorG}{HTML}{000000}
\definecolor{colorH}{HTML}{777777}
\definecolor{colorI}{HTML}{bdd6ff}
\definecolor{colorJ}{HTML}{6a9e6f}

\usepackage{pifont}

\newcommand{\expnum}[2]{$#1\!\times\!10^{#2}$}

\begin{document}

\begin{acronym}
\acro{AD}{automatic differentiation}
\acro{CDC}{CLIP direction consistency}
\acro{DoG}{difference of Gaussian}
\acro{MLP}{multi-layer perceptron}
\acro{NeRF}{Neural Radiance Field}
\acro{NNF}{nearest neighbour field}
\acro{NVS}{novel view synthesis}
\acro{PSNR}{peak signal-to-noise ratio}
\acro{SDF}{signed distance field}
\acro{SSIM}{Structural Similarity Index Measure}
\acro{ViT}{vision transformer}
\end{acronym}

\newcommand{\task}[1]{\textsc{#1}}
\newcommand{\method}[1]{{\texttt{#1}}}
\newcommand{\mypara}[1]{\noindent\textbf{#1:}\quad}

\renewcommand{\eg}{\textit{e.g.}, }
\renewcommand{\ie}{\textit{i.e.}, }
\renewcommand{\wrt}{w.r.t.\ }

\newcommand{\noval}{\hl{0.0}}

\title{NeRF Analogies: Example-Based Visual Attribute Transfer for NeRFs}

\author{%
Michael Fischer\textsuperscript{1}\thanks{Corresponding author. Work done during an internship at Meta Reality Labs Research. Contact: m.fischer@cs.ucl.ac.uk.} \quad
Zhengqin Li\textsuperscript{2} \quad
Thu Nguyen-Phuoc\textsuperscript{2} \quad
Alja\v{z} Bo\v{z}i\v{c}\textsuperscript{2} \quad
Zhao Dong\textsuperscript{2} \\[0.75ex]
Carl Marshall\textsuperscript{2} \quad
Tobias Ritschel\textsuperscript{1} \\[1.0ex]
\textsuperscript{1}University College London \quad
\textsuperscript{2}Meta Reality Labs Research
}

\makeatletter
\g@addto@macro\@maketitle{
  \begin{figure}[H]
  \vspace{-0.75cm}
  \setlength{\linewidth}{\textwidth}
  \setlength{\hsize}{\textwidth}
  \centering
  \includegraphics[width=\textwidth]{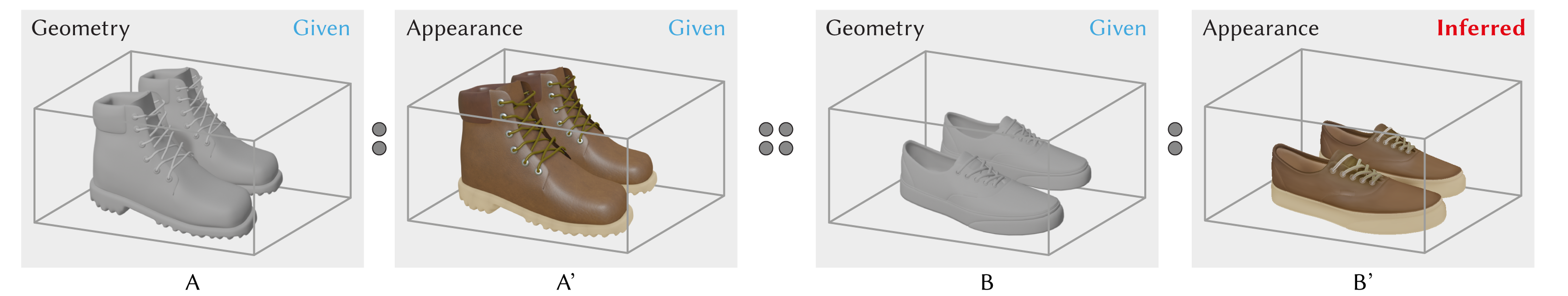}
  \vspace{-0.5cm}
  \caption{Given a source NeRF that encodes 3D geometry (A) and appearance (A'), as well as a target NeRF that encodes 3D geometry without appearance (B), our method infers a \emph{NeRF analogy} (B') that combines the target geometry with the source appearance.}
  \end{figure}
  \label{fig:Teaser}
}
\makeatother
\maketitle

\mymath{\appearances}{L}
\mymath{\edgeloss}{\mathcal{L}_G}
\mymath{\features}{\mathbf f}
\mymath{\normals}{\mathbf n}
\mymath{\featureImage}{\mathcal F}
\mymath{\gaussianOne}{G_{\sigma_1}}
\mymath{\gaussianTwo}{G_{\sigma_2}}
\mymath{\image}{\mathcal I}
\mymath{\lossweight}{\lambda}
\mymath{\mapping}{\phi}
\mymath{\parameters}{\theta}
\mymath{\positions}{\mathbf x}
\mymath{\radiance}{\mathcal R}
\mymath{\representation}{L_\parameters}
\mymath{\viewDirections}{\omega}
\mymath{\currentImage}{\image^\mathrm{Current}}
\mymath{\sourceRadiance}{\radiance^\mathrm{Source}}
\mymath{\targetRadiance}{\radiance^\mathrm{Target}}
\mymath{\sourceImage}{\image^\mathrm{Source}}
\mymath{\targetImage}{\image^\mathrm{Target}}
\mymath{\sourceFeatureImage}{\featureImage^\mathrm{Source}}
\mymath{\targetFeatureImage}{\featureImage^\mathrm{Target}}
\mymath{\sourcePointCount}n
\mymath{\targetPointCount}m
\mymath{\sourceFeatures}{\features^\mathrm{Source}}
\mymath{\targetFeatures}{\features^\mathrm{Target}}
\mymath{\sourceAppearances}{\appearances^\mathrm{Source}}
\mymath{\targetAppearances}{\appearances^\mathrm{Target}}
\mymath{\sourceViewDirections}{\viewDirections^\mathrm{Source}}
\mymath{\targetViewDirections}{\viewDirections^\mathrm{Target}}
\mymath{\targetNormals}{\normals^\mathrm{Target}}
\mymath{\targetPositions}{\positions^\mathrm{Target}}
\mymath{\sourceIndex}{i}
\mymath{\targetIndex}{j}
\mymath{\similarity}{p}

\begin{abstract}
A \ac{NeRF} encodes the specific relation of 3D geometry and appearance of a scene.
We here ask the question whether we can transfer the appearance from a source \ac{NeRF} onto a target 3D geometry in a semantically meaningful way, such that the resulting new \ac{NeRF} retains the target geometry but has an appearance that is an analogy to the source \ac{NeRF}.
To this end, we generalize classic image analogies from 2D images to \acp{NeRF}. 
We leverage correspondence transfer along semantic affinity that is driven by semantic features from large, pre-trained 2D image models to achieve multi-view consistent appearance transfer.
Our method allows exploring the mix-and-match product space of 3D geometry and appearance.
We show that our method outperforms traditional stylization-based methods and that a large majority of users prefer our method over several typical baselines. 
Project page: \url{mfischer-ucl.github.io/nerf\_analogies}.
\end{abstract}

\mysection{Introduction}{Introduction}
Understanding and representing the 
three-dimensional world, a fundamental challenge in computer vision, requires accurate modeling of the interplay between geometry and appearance. 
\ac{NeRF}s \cite{mildenhall2021nerf} have emerged as a pivotal tool in this space, uniquely encoding this relationship via optimized color- and density-mappings. 
However, in spite of their success for high-quality \ac{NVS}, most \ac{NeRF} representations remain notoriously hard to edit, which led to the research field of \ac{NeRF} editing.

In this work, we contribute to this evolving landscape by exploring \emph{\ac{NeRF} Analogies}, a novel aspect of \ac{NeRF} manipulation between semantically related objects. 
Fig.~1 illustrates this concept: We begin with an existing \ac{NeRF}, designated $A'$, which is derived from the geometric structure of a boot ($A$) and its appearance.
This \ac{NeRF}, which we henceforth will call our \emph{source} \ac{NeRF}, encodes the relation of geometry and appearance.
In a \emph{\ac{NeRF} analogy}, we now seek to infer a new \ac{NeRF}, $B'$, which, given a \emph{target} geometric shape (the sneaker, $B$), satisfies the analogy \emph{$A:A' :: B:B'$}, \ie combines the visual appearance of $A'$ with the new geometry $B$.
\ac{NeRF} analogies hence are a way of changing a \ac{NeRF}'s geometry while maintaining its visual appearance - a counterpoint to recent research, which often aims to update the encoded appearance based on (potentially non-intuitive) text-embeddings, while keeping the geometry (largely) unchanged.

Creating a \ac{NeRF} analogy essentially requires solving the problem of finding semantically related regions between the target geometry $B$ and the existing source \ac{NeRF} $A'$, which will then serve as guides for the subsequent appearance transfer.
While image-based correspondence has been thoroughly researched in the past \cite{ng2003sift, bay2008speeded, rublee2011orb, lowe2004distinctive, fischer2023plateau, fischler1981random, shi1994good}, recent work has shown the (un)reasonable success of the activations of large, pre-trained networks for the task of (dense) \emph{semantic} image correspondence \cite{sharma2023materialistic, amir2021deep, bhattad2023stylegan, zhang2023tale}. 
More specifically, \citet{amir2021deep} and \citet{sharma2023materialistic} both show that the features produced by the attention layers in \acp{ViT} can be used as expressive descriptors for dense semantic correspondence tasks, presumably due to the attention mechanism's global context \cite{amir2022effectiveness, hedlin2023unsupervised, luo2023diffusion, tang2023emergent}. 

In this work, we thus leverage the expressiveness of DiNO-ViT, a large pre-trained vision transformer \cite{caron2021emerging} to help us generalize classic Image Analogies \cite{hertzmann2023image} from two-dimensional images to multiview-consistent light fields. 
To this end, we compute the semantic affinity between pixel-queries on renderings of the 3D target geometry and 2D slices of the source \ac{NeRF} via the cosine-similarity of the produced \ac{ViT} features, and subsequently use this mapping to transfer the visual appearance from the source onto the target.
Assuming we can query the 3D position of our target geometry, repeating this process over many views and pixels results in a large corpus of position-appearance pairs, which we use as input for training our \ac{NeRF} analogy $B'$, thereby achieving a multiview-consistent 3D representation that combines target geometry and source appearance.

We compare NeRF analogies to other methods via quantitative evaluation and a user-study and find that a significant majority of users prefer our method.
NeRF analogies allow exploring the product space of 3D geometry and appearance and provide a practical way of changing neural radiance fields to new geometry while keeping their original appearance. 

\mycfigure{Overview}{
The main steps of our approach from left to right:
We render both the target and source \ac{NeRF} (first and second pair of rows) into a set of 2D images (first column), and then extract features (middle column).
An image hence is a point cloud in feature space, where every point is labeled by 3D position, normal, view direction and appearance (third column).
We use view direction, RGB and features of the source \ac{NeRF}, and position, normal and features of the target, and gray-out unused channels.
We then establish correspondence between the source and target features via the mapping \mapping in the lower right subplot, allowing us to transfer appearance from the source to the geometry of the target. 
Finally, we train our \ac{NeRF} analogy \representation which combines the target's geometry with the appearance from the source.
}

\mysection{Previous Work}{PreviousWork}
\textbf{Example-based editing} so far has largely been done in 2D, \eg via the seminal PatchMatch algorithm \cite{barnes2009patchmatch}, image analogies \cite{hertzmann2023image}, deep image analogies \cite{liao2017visual}, style transfer \cite{gatys2015neural}, example based visual attribute transfer \cite{fivser2016stylit, he2019progressive, rematas2014image} or, most recently, through \ac{ViT}- or diffusion-features \cite{vsubrtova2023diffusion, tumanyan2022splicing}. 
Here, a reference (source) image or style is provided and used to update a content image (the target). 
These techniques have been shown to work well and to be intuitive, as the user can intuitively control the outcome by changing the style image (\ie there is no black-box, like the prompt-embedding in text-based methods), but are limited to 2D. 
Most cannot easily be lifted to 3D (\ie by multiview-training and backpropagation to a common underlying representation), as many of the employed operations are non-differentiable (\eg the \ac{NNF} search in \cite{hertzmann2023image} or up-scaling by res-block inversion in \cite{liao2017visual}).
Hence, when they are na\"\i vely lifted to 3D by training a \ac{NeRF} on the 2D output, the result will be of low quality, as different output views are not consistent, leading to floaters and density artifacts in free space. 

\textbf{Neural Radiance Fields} \cite{mildenhall2021nerf, muller2022instant, barron2021mip, barron2022mip} do not have this problem, as they solve for an underlying 3D representation during multiview-training, \ie the output is enforced to be consistent by simultaneously training on multiple views of the scene.
However, editing \ac{NeRF}s is a notoriously hard problem, as often geometry and appearance are entangled in a non-trivial way and non-intuitive, implicit representation. 

\textbf{\ac{NeRF} editing} hence often either simplifies this by separate editing of shape \cite{yuan2022nerf, liu2022devrf, xu2022deforming, jambon2023nerfshop, chen2023neuraleditor} or appearance \cite{kuang2023palettenerf, zhang2023dyn, wu2023nerf, wang2023proteusnerf}, recently also text-based \cite{song2023blending, wang2022clip, haque2023instruct}. 
Another branch of work is the stylization of \ac{NeRF}s \cite{nguyen2022snerf, liu2023stylerf, zhang2022arf, huang2021learning, huang2022stylizednerf, wang2023nerf}, which uses methods from neural style transfer \cite{gatys2015neural} to stylize the underlying \ac{NeRF}, either via stylizing the captured images or through stylization of 3D feature volumes. 
Most of the aforementioned methods, however, ignore semantic similarity while performing stylization or appearance editing, with the exception of \cite{kobayashi2022decomposing, pang2023locally, bao2023sine}, who perform region-based stylization or appearance-editing of \acp{NeRF}, but do not change geometry.
For an overview of the vast field of \acp{NeRF} and their editing techniques, we refer to the excellent surveys \cite{gao2022nerf} and \cite{tewari2022advances}.

\textbf{Limitations} of many of the above approaches include that they are often solving for \emph{either} shape or appearance changes, and that the recently popular text-embeddings often might not produce the exact intended result (we show an example in \refFig{instructn2n}).
Moreover, many \ac{NeRF} shape editing methods are restricted to small or partial shape changes, as they solve for a deformation field and thus are restricted to a limited amount of change (\eg excluding topological changes \cite{bao2023sine, kobayashi2022decomposing, xu2022deforming}).
We aim to make progress in the direction of \emph{combined and multiview-consistent} semantic appearance-editing by introducing \emph{NeRF analogies}, combining target geometry with a source appearance. 

\textbf{Inter-Surface Mappings}, in pursuit of a similar goal, try to establish relations between two shapes by comparing their geometric \cite{eisenberger2020smooth, schmidt2023surface} or, more recently, semantic \cite{morreale2023neural, abdelreheem2023zero} features.
However, most surface mapping methods either rely on manual annotations (\ie are non-automatic) \cite{schmidt2023surface}, are non-robust to geometry imperfections \cite{eisenberger2020smooth}, introduce discontinuous partitions \cite{abdelreheem2023zero, morreale2021neural} or are limited to objects of the same topology (\eg genus-zero surfaces \cite{morreale2023neural}) and hence are currently not suitable for the task of \ac{NeRF} analogies, but might provide an interesting direction for future research.

\mysection{Our Approach}{OurApproach}

The following sections will first formalize the abstract idea (\refSec{Idea}) and subsequently describe our specific implementation (\refSec{Implementation}).
\mysubsection{\ac{NeRF} Analogies}{Idea}

\paragraph{Feature extraction}
As mentioned previously, the source (radiance) \ac{NeRF} \sourceRadiance provides view-dependent RGB color, while the target \ac{NeRF} \targetRadiance provides geometry. 
Rendering \sourceRadiance and \targetRadiance from a range of random view directions produces the first three rows of the first column in \refFig{Overview}, while the fourth row in that column is the result we aim to compute.

We then use these renderings to compute dense feature descriptors of all images (visualized as the false-color images in the second column of \refFig{Overview}).
We require this feature embedding to place semantically similar image parts in close regions of the embedding space.

\myfigure{affinity_visualized}{DiNO affinity for various pixel queries (colored dots, columns) on various object pairs (rows), visualized as heatmap where blue and red correspond to 0 and 1, respectively.}

For all renderings, we store the per-pixel features, the RGB color, the 3D position and the viewing directions of all non-background pixels into two large vectors, \sourceFeatureImage and \targetFeatureImage.
These are best imagined as point clouds in feature space, where some points are labeled as appearance and others as view direction, as seen in the last column of \refFig{Overview}.
This pair of point clouds will serve as supervision for our training, which will be explained next.
The figure also shows grayed-out what is not relevant: positions of the source and the RGB and view direction of the target.

\paragraph{Training}
In order to combine the appearance of the source with the geometry of the target, we train a 3D-consistent \ac{NeRF} representation on the previously extracted point clouds \sourceFeatureImage and \targetFeatureImage. 
As the target geometry is given, we only need to learn the view-dependent appearance part of that field.
With a given geometry, and the appearance given at 3D coordinate, this a simple direct supervised learning, that does not even require differentiable rendering.
The key challenge is, however, to identify where, and under which viewing angle, the relevant appearance information for a given target location is to be found in the source.

To this end, we sample \sourcePointCount locations in \sourceFeatureImage (shown as red dots in \refFig{Overview}), and, at each location, extract the source feature descriptors \sourceFeatures, the source appearance \sourceAppearances, and the source viewing directions \sourceViewDirections.
Similarly, we also sample \targetPointCount locations from the target point cloud \targetFeatureImage (shown as blue dots in \refFig{Overview}) and, at each location, fetch
the image features \targetFeatures and
the target positions \targetPositions.

Now, we find a discrete mapping 
$
\mapping_\targetIndex
\in
(1,\ldots,\targetPointCount)
\rightarrow
(1,\ldots,\sourcePointCount)
$
that maps every target location index \targetIndex to the source location index \sourceIndex with maximal similarity 
:
\[
\mapping_\targetIndex
:=
\operatorname{arg\,max}_\sourceIndex
\operatorname{sim}(
\targetFeatures_\targetIndex,
\sourceFeatures_\sourceIndex
)
.
\]
As $\targetPointCount\times\sourcePointCount$ is a moderate number, this operation can be performed by constructing the full matrix, parallelized across the GPU, and finding the maximal column index for each row.
The mapping \mapping is visualized as the links between nearby points in the overlapping feature point clouds in the lower right in \refFig{Overview}.
Notably, we do not enforce \mapping to be 3D-consistent or bijective, as this would constrain the possible transfer options (consider the case where the appearance transfer would need to be a 1:n mapping, \eg when transferring the appearance from a single-legged chair onto a four-legged table). 
Instead, we ask for the feature with the maximum similarity and rely on the feature extractor to find the correct color-consistent matches across multiple views.

Now, define
$
\targetAppearances_\targetIndex = 
\sourceAppearances_{\mapping_\targetIndex}
$
as the appearance that the target should have under the mapping \mapping and a certain viewing direction, given the extracted correspondences. 

This information -- i) position, ii)  direction, and iii) radiance -- is commonly sufficient to train the appearance part of a radiance field:
i) The target 3D positions are known, as they can be directly inferred from the target geometry and its renderings.
ii) The source view direction is known on the source, and we would like the target's view-dependence to behave the same.
iii) The appearance is known from the source via the mapping \mapping.
Notably, i) implies that the density function decays to a Dirac delta distribution, so no volume rendering is required - the appearance values simply have to be correct at the correct positions in space.
Moreover, we found it beneficial to add the target's surface normal into the network to provide high-frequency input signal that aids in recovering high-frequent color changes. 

We thus train the parameters \parameters of our NeRF Analogy $\appearances_\theta$ (for network details see Suppl. Sec. 1) such that for every observed target position, target and source appearance match under the source viewing direction, \ie 
\vspace{0.1cm}
\[
\label{eq:colorloss}
\mathbb E_\targetIndex
[
|
\representation(\targetPositions_\targetIndex, \targetNormals_\targetIndex, \targetViewDirections_\targetIndex)
- \mapping_\targetIndex(\sourceAppearances_\sourceIndex, \sourceViewDirections_\sourceIndex)
|_1
].
\]
\vspace{-0.2cm}
\myfigure{multires_affinity}{Self-similarity for a pixel query (the yellow point on the left image) for several variants of DiNO to illustrate the effects of feature resolution. 
Our version produces the most fine-granular features, as is visible in the rightmost image.}
\vspace{-0.2cm}
\mysubsection{Implementation}{Implementation}
\paragraph{Features}
Expressive features are crucial for establishing correspondences between objects and their semantic parts. 
We rely on the features produced by DiNO-\ac{ViT} \cite{caron2021emerging}, which have been shown to capture both semantic and structural information to a high extent \cite{amir2021deep, sharma2023materialistic} and defer discussion of our exact ViT setup to the supplemental for brevity.

\myfigure{pca}{
Visualization of the first three PCA components of the features computed across both images as RGB colors.
Semantically similar regions have similar descriptors, hence similar colors.
Transferring the appearance along the most similar descriptor for each pixel creates the middle image.
}

\refFig{multires_affinity} shows the comparison between the original ViT granularity, \citet{amir2021deep}'s reduced strides and our feature granularity, while \refFig{pca} visualizes the semantic correspondence via the first three principal components of the ViT features computes between the two images.
As commonly done, we compare the features according to their cosine similarity
\[
\operatorname{sim}(\features_1,\features_2)
:=
\frac{
\left<
\features_1,
\features_2
\right>
}
{
||\features_1||
\cdot
||\features_2||
}
.
\]

As per the previous explanations in \refSec{Idea}, the general idea behind our approach does not need, and never makes use of, the target geometry's color or texture. 
However, as our feature extractor DiNO was trained on natural images, we found its performance to decrease on un-textured images and thus use textured geometry. 
We show an ablation of this in \refSec{Ablation} and are confident that future, even more descriptive feature extractors will be able to match correspondence quality on untextured meshes. 

\paragraph{Sampling}
We randomly render 100 images per object. 
From each image, we sample an empirically determined number of 5,000 non-background pixels to compute their features and affinities.
For each of these sampled pixels, we need to compute the cosine similarity to all feature descriptors in the source \ac{NeRF}. 
In practice, we employ importance-sampling for the cosine similarity and constrain the similarity computation to the feature descriptors of the 10 closest views.
While this approach might introduce slight bias, we found it to work well in practice (as we are only interested in the maximum similarity, and similar views generally will have higher similarity scores) while significantly reducing computation time. 
As many image analogy and appearance transfer frameworks, we assume roughly aligned objects \cite{hertzmann2023image, liao2017visual,luo2023diffusion,fischer2022metappearance}, \ie similar orientation and pose. 
For non-aligned objects, we run a pre-conditioning step by optimizing rotation and translation such that the objects roughly align.

\paragraph{Edge loss}
As DiNO-ViT is a 2D method that we employ in a 3D context, it is inevitable that some of the feature correspondences will be noisy across different views, \ie we cannot guarantee that a certain image part will map to the same location under a different view. In our training setup, this leads to washed-out details, which are especially notable in high-frequency regions, \eg around edges, and on the silhouettes of the objects.  
We alleviate this by computing an additional regularization term that enforces the \acp{DoG} between monochrome versions of the current rendering \currentImage and the target image \targetImage to coincide:
\[
    \edgeloss = |\,\currentImage * \gaussianOne - \targetImage * \gaussianTwo\,|_1
\]
where $\ast$ denotes convolution. 
We use standard deviations $\sigma_1 = 1.0$ and $\sigma_2 = 1.6$, which is a common choice for this filter \cite{marr1980theory}. 
We add this term to our training loss in order to bring back edge detail, weighted by a factor \lossweight, which we set to zero during the first 15\% of the training in order to allow the network to learn the correct colors first before gradually increasing it to an empirically determined value of 50.
We show an ablation of this loss in \refSec{Ablation} and detail further training details in the supplemental. 

\mycfigure{results}{Results of our method and related approaches on object pairs. For every row, we use the target's geometry (top row) and the appearance of the source \ac{NeRF} (bottom row). The multiview-consistency of these results is best appreciated in the supplemental videos.}

\mycfigure{real_results}{Results on the real-world \textsc{kitchen} (top), \textsc{garden} (middle) and \textsc{truck} scenes from the MiP-NeRF 360 \cite{barron2022mip} and Tanks and Temples \cite{knapitsch2017tanks} datasets, respectively. For each experiment, we show the original scene in the leftmost subplot, followed by the target geometry, our inferred NeRF analogy and the baselines style transfer \cite{gatys2015neural}, WCT \cite{li2017universal} and Deep Image Analogies \cite{liao2017visual}.}
\mysection{Results}{Results}
As we are, to the best of our knowledge, the first to introduce semantically meaningful appearance transfer onto arbitrary 3D geometry, there are no directly applicable comparisons to evaluate. 
Nonetheless, we compare to traditional image-analogy and style-transfer methods such as Neural Style Transfer \cite{gatys2015neural},  WCT \cite{li2017universal} and Deep Image Analogies \cite{liao2017visual} by running them on pairs of images and then training a \ac{NeRF} (we use InstantNGP \cite{muller2022instant}) on the resulting output images.
For style transfer, WCT and deep image analogies, we use the publicly available implementations \cite{pytorchstyletransfer}, \cite{wctpytorch} and \cite{deepimageanalogiespytorch}, respectively. 
Those methods will necessarily produce floaters and free-space density artifacts, as they are not multiview-consistent. 
To allow a fairer comparison, we reduce these artifacts by multiplying their output with the target's alpha-channel.
Moreover, we compare to the 3D-consistent appearance transfer method S\ac{NeRF} \cite{nguyen2022snerf}, which runs style transfer in conjunction with \ac{NeRF} training and whose authors kindly agreed to run their method on our data. 
In accordance with the stylization and image-analogy literature, we use the target as content- and the source as style-image.

\paragraph{Qualitative}
We show results of our method and its competitors on various object pairs in \refFig{results}. 
It becomes evident that style-based methods fail to produce sharp results and do not capture semantic similarity (\eg the bag's handle is not brown, the chair's legs are not beige). 
Deep Image Analogies (DIA, \cite{liao2017visual}) manages to achieve crisp details, as it stitches the output together from pieces of the input, but does not capture the target's details well (cf. the green chair's backrest or the boots' laces).
As is seen from the videos in the supplemental material, none of the methods except S\ac{NeRF} and ours are multiview-consistent, which leads to floaters, artifacts and inconsistent color changes. 

We further show a \ac{NeRF} analogy on a challenging multi-object scene in \refFig{tables}. 
The challenge here arises from object-level ambiguities (no unique mapping between the two table tops), semantic gaps (sofas on the left vs.\ chairs on the right) and many-to-many relation (2 sofas vs. 4 chairs).
In spite of not being perfect (\eg the couch appearance bleeding onto parts of the table edges), our method handles this case well and transfers the appearance among semantically related objects (\eg apples, plants, chairs).

\myfigure{instructn2n}{Text-based methods often cannot accurately represent the desired geometry, or the editing fails completely, as seen here. For our results on these scenes and views, see \refFig{real_results}.}

Finally, we show results on real-world scenes from the MiP-NeRF 360 \cite{barron2022mip} and Tanks and Temples \cite{knapitsch2017tanks} datasets in \refFig{real_results}. 
We replace parts of the encoded geometry by first manually finding a bounding box and then setting the volume density withing that box to zero. 
Rendering now results in the object being cut-out of the scene while inverting the box results in a rendering of the object only. 
Those renderings constitute our source appearance and, in conjunction with a provided target geometry, allow us to create a \ac{NeRF} analogy which we can composit back into the original source \ac{NeRF} via the painter's algorithm. 
As \refFig{real_results} shows, our method produces consistent results and transfers the appearance in a semantically meaningful way. 
Interestingly, the comparison with the state-of-the-art text-based \ac{NeRF} editing method Instruct-Nerf2Nerf  \cite{haque2023instruct} in \refFig{instructn2n} shows that their model cannot capture the required level of detail and thus fails to perform the required edits - a limitation of the underlying InstructPix2Pix's text-embedding \cite{brooks2023instructpix2pix}.

\paragraph{Quantitative}
The popular image metrics \ac{PSNR} and \ac{SSIM} require ground-truth reference images. For the case of \ac{NeRF} analogies, these do not exist, as the ''best`` semantic mapping is subjective and therefore cannot be uniquely determined. 
We therefore report ``bootstrapped'' \ac{PSNR} and \ac{SSIM} (BPSNR, BSSIM) values as follows: 
We first create a \ac{NeRF} analogy, and then use it to render out all train and test images. 
The rendered-out train images then serve as input for training a new \ac{NeRF} (we use Instant-NGP \cite{muller2022instant}), from which we compute \ac{PSNR} and \ac{SSIM} between the previously- and now newly rendered test images. 
While these metrics must nominally not be compared with values reported by traditional \ac{NeRF} methods (we are testing on the output of a method, not on ground-truth data), they serve as a relative (\ie inter-method) indicator of multiview-consistency: if the rendered-out test images are multiview-consistent, the reconstruction will produce similar images, resulting in higher BPSNR and BSSIM values.
Inspired by \cite{haque2023instruct}, we additionally report values for the \ac{CDC} (cf. the supplemental for more details). 

We detail the quantitative results of the aforementioned metrics in \refTab{results} and find that they support the qualitative results: our method achieves the highest score across all metrics, while style transfer and WCT achieve the lowest scores. 
S\ac{NeRF} produces better results but still shows relatively low scores, presumably because its style-transfer module struggles with the type of styles used in our experiments (the source appearance is highly non-stationary, which is known to cause problems to VGG-based methods). 
DIA performs surprisingly well on most metrics, although it does not have 3D information. 
We attribute this to the fact that it creates its output from patching together parts of the input, where the input by design is 3D consistent. 

\newcolumntype{P}{r<{\%}}

\begin{table}[htb]
\small
\centering 
\renewcommand{\tabcolsep}{0.092cm}
\caption{Quantitative results of our and other methods according to different metrics (cf. \refSec{Results} for details). Higher is better for all, the best and second-best results are bold and underlined, respectively.}
\begin{tabular}{lrrrPPPP}
\toprule 
&
\multicolumn3c{Metrics}&
\multicolumn4c{User study}
\\
\cmidrule(lr){2-4}
\cmidrule(lr){5-8}
&
\multicolumn1c{\scriptsize BPSNR}&
\multicolumn1c{\scriptsize BSSIM} &
\multicolumn1c{\scriptsize CLIP} &
\multicolumn1c{\scriptsize Transfer} &
\multicolumn1c{\scriptsize MVC}  &
\multicolumn1c{\scriptsize Quality} & 
\multicolumn1c{\scriptsize Comb.} \\ \midrule
ST \cite{gatys2015neural} & 25.14 & .870 & .981  & 1.7 & 1.4 & 2.9 & 1.9 \\
WCT \cite{li2017universal} & 28.64 & .917 & .983 & 3.4 & 0.5 & 0.5 & 1.9 \\
DIA \cite{liao2017visual} & \underline{33.06} & \underline{.968} & .983  & \underline{28.6} & \underline{20.5} & \underline{9.1} & \underline{23.0}\\
S\ac{NeRF} \cite{nguyen2022snerf} & 32.41 & .947 & \underline{.984} & 7.8 & 1.0 & 2.9 & 4.8\\
Ours & \textbf{36.16} & \textbf{.984} & \textbf{.992} & \textbf{58.5} & \textbf{76.7} & \textbf{84.8} & \textbf{68.4} \\
\bottomrule
\end{tabular}
\label{tab:results}
\end{table}

\myfigure{tables}{A \ac{NeRF} analogy on a multi-object scene. The semantic mapping correctly matches apples, plants, tables and chairs.}

\mysubsection{User Study}{Userstudy}
In addition to the quantitative evaluation previously described, we ran a user study to complement the evaluation and to assess our method's semantic quality, \ie whether the produced output is a plausible mixture of target geometry and source appearance.
In the first section of the study, participants were shown 2D results in randomized order and asked which approach best combines the target geometry and the source appearance (``Transfer'' in \refTab{results}). 
For the second study, we lifted the 2D methods to 3D by using their outputs as input images for InstantNGP \cite{muller2022instant}.
We then rendered a circular camera trajectory and asked the participants to choose their preferred method for a) multi-view consistency and b) floaters and artifacts (``MVC'' and ``Quality'' in \refTab{results}, respectively). 
We gathered responses from 42 participants and show the averaged responses across all three categories in the rightmost column (``Combined'') of \refTab{results}.
The results in \refTab{results} support our quantitative and qualitative findings and show our method to be leading by a wide margin across all categories. 
All statements are highly significant with a Pearson $\xi$-square test's $p<0.001$.

\myfigure{rankings}{Outcome of our user study, as also per \refTab{results}.}

\mysubsection{Ablation Study}{Ablation}
We show each of our design decision's influence with an ablation, displayed qualitatively in \refFig{ablation}. 
Ablating the edge-loss still allows clear color distinctions between semantic parts (\eg front vs.\ roof), but leads to lower-frequent detail, as the network has no incentive to learn the target's fine-granular details. 
While ablating the texture on the target geometry leads to slightly washed-out descriptors, presumably due to more noise in the DiNO-ViT correspondences, our method still produces semantically similar results, supporting our claim that DiNO features are expressive and translate across the domain gap between textured and untextured geometry. 

\myfigure{ablation}{Ablation of parts of our method: ablating the edge-loss leads to loss of detail, while ablating the target's texture leads to noisier DiNO correspondences and hence lower-quality results.}

\mysection{Limitations}{Limitations}
Our method fundamentally relies on the mapping \mapping and the \ac{NeRF} representation \radiance. 
While \ac{NeRF} analogies harness the strengths of these methods, they also align with their unique limitations.
It is, for instance, hard for DiNO (and most other correspondence methods) to resolve rotational ambiguities on round objects.
Moreover, as we perform point-based appearance transfer, we are unable to transfer texture. 
We show a failure case in \refFig{failure}.

\myfigure{failure}{A limitation of our method becomes apparent on this challenging case, where, due to specularity and rotation-symmetry, the DiNO correspondences are inaccurate.
The middle column shows that our method tries to consolidate these inaccuracies by erroneously encoding the different colors in the viewing directions}
\vspace{-0.2cm}
\mysection{Conclusion}{Conclusion}
In this work, we have introduced \emph{\ac{NeRF} analogies}, a framework for visual attribute transfer between \ac{NeRF}s via semantic affinity from \ac{ViT} features. 
Our method can be used to aid in content-creation, \eg by combining user-captured geometry with appearance from online 3D models, and also works in multi-object settings and on real-world scenes.
We compare favorably against other methods from the color-transfer, image synthesis and stylization literature and achieve the highest rankings in a user study both for transfer quality and multiview-consistency.
\ac{NeRF} analogies can open up exciting areas of future research, such as 3D-consistent texture transfer or the transfer of intrinsic scene parameters such as roughness or specular albedo.
Another interesting research direction could be to learn the sampling pattern to find the most relevant directions or views \cite{liu2023learning, kopanas2023improving, lee2023so} for the subsequent learning of a NeRF analogy.

\setlength{\bibsep}{0.0pt}
{\small
\bibliographystyle{plainnat}
\bibliography{main}

\begin{thebibliography}{72}
\providecommand{\natexlab}[1]{#1}
\providecommand{\url}[1]{\texttt{#1}}
\expandafter\ifx\csname urlstyle\endcsname\relax
  \providecommand{\doi}[1]{doi: #1}\else
  \providecommand{\doi}{doi: \begingroup \urlstyle{rm}\Url}\fi

\bibitem[Abdelreheem et~al.(2023)Abdelreheem, Eldesokey, Ovsjanikov, and Wonka]{abdelreheem2023zero}
Ahmed Abdelreheem, Abdelrahman Eldesokey, Maks Ovsjanikov, and Peter Wonka.
\newblock Zero-shot 3d shape correspondence.
\newblock \emph{arXiv preprint arXiv:2306.03253}, 2023.

\bibitem[Amir et~al.(2021)Amir, Gandelsman, Bagon, and Dekel]{amir2021deep}
Shir Amir, Yossi Gandelsman, Shai Bagon, and Tali Dekel.
\newblock Deep vit features as dense visual descriptors.
\newblock \emph{arXiv preprint arXiv:2112.05814}, 2\penalty0 (3):\penalty0 4, 2021.

\bibitem[Amir et~al.(2022)Amir, Gandelsman, Bagon, and Dekel]{amir2022effectiveness}
Shir Amir, Yossi Gandelsman, Shai Bagon, and Tali Dekel.
\newblock On the effectiveness of vit features as local semantic descriptors.
\newblock In \emph{European Conference on Computer Vision}, pages 39--55. Springer, 2022.

\bibitem[Bao et~al.(2023)Bao, Zhang, Yang, Fan, Yang, Bao, Zhang, and Cui]{bao2023sine}
Chong Bao, Yinda Zhang, Bangbang Yang, Tianxing Fan, Zesong Yang, Hujun Bao, Guofeng Zhang, and Zhaopeng Cui.
\newblock Sine: Semantic-driven image-based nerf editing with prior-guided editing field.
\newblock In \emph{The IEEE/CVF Computer Vision and Pattern Recognition Conference (CVPR)}, 2023.

\bibitem[Barnes et~al.(2009)Barnes, Shechtman, Finkelstein, and Goldman]{barnes2009patchmatch}
Connelly Barnes, Eli Shechtman, Adam Finkelstein, and Dan~B Goldman.
\newblock Patchmatch: A randomized correspondence algorithm for structural image editing.
\newblock \emph{ACM Trans. Graph.}, 28\penalty0 (3):\penalty0 24, 2009.

\bibitem[Barron et~al.(2021)Barron, Mildenhall, Tancik, Hedman, Martin-Brualla, and Srinivasan]{barron2021mip}
Jonathan~T Barron, Ben Mildenhall, Matthew Tancik, Peter Hedman, Ricardo Martin-Brualla, and Pratul~P Srinivasan.
\newblock Mip-nerf: A multiscale representation for anti-aliasing neural radiance fields.
\newblock In \emph{Proceedings of the IEEE/CVF International Conference on Computer Vision}, pages 5855--5864, 2021.

\bibitem[Barron et~al.(2022)Barron, Mildenhall, Verbin, Srinivasan, and Hedman]{barron2022mip}
Jonathan~T Barron, Ben Mildenhall, Dor Verbin, Pratul~P Srinivasan, and Peter Hedman.
\newblock Mip-nerf 360: Unbounded anti-aliased neural radiance fields.
\newblock In \emph{Proceedings of the IEEE/CVF Conference on Computer Vision and Pattern Recognition}, pages 5470--5479, 2022.

\bibitem[Bay et~al.(2008)Bay, Ess, Tuytelaars, and Van~Gool]{bay2008speeded}
Herbert Bay, Andreas Ess, Tinne Tuytelaars, and Luc Van~Gool.
\newblock Speeded-up robust features (surf).
\newblock \emph{Computer vision and image understanding}, 110\penalty0 (3):\penalty0 346--359, 2008.

\bibitem[Bhattad et~al.(2023)Bhattad, McKee, Hoiem, and Forsyth]{bhattad2023stylegan}
Anand Bhattad, Daniel McKee, Derek Hoiem, and DA~Forsyth.
\newblock Stylegan knows normal, depth, albedo, and more.
\newblock \emph{arXiv preprint arXiv:2306.00987}, 2023.

\bibitem[Brooks et~al.(2023)Brooks, Holynski, and Efros]{brooks2023instructpix2pix}
Tim Brooks, Aleksander Holynski, and Alexei~A Efros.
\newblock Instructpix2pix: Learning to follow image editing instructions.
\newblock In \emph{Proceedings of the IEEE/CVF Conference on Computer Vision and Pattern Recognition}, pages 18392--18402, 2023.

\bibitem[Caron et~al.(2021)Caron, Touvron, Misra, J{\'e}gou, Mairal, Bojanowski, and Joulin]{caron2021emerging}
Mathilde Caron, Hugo Touvron, Ishan Misra, Herv{\'e} J{\'e}gou, Julien Mairal, Piotr Bojanowski, and Armand Joulin.
\newblock Emerging properties in self-supervised vision transformers.
\newblock In \emph{Proceedings of the IEEE/CVF international conference on computer vision}, pages 9650--9660, 2021.

\bibitem[Chen et~al.(2023)Chen, Lyu, and Wang]{chen2023neuraleditor}
Jun-Kun Chen, Jipeng Lyu, and Yu-Xiong Wang.
\newblock Neuraleditor: Editing neural radiance fields via manipulating point clouds.
\newblock In \emph{Proceedings of the IEEE/CVF Conference on Computer Vision and Pattern Recognition}, pages 12439--12448, 2023.

\bibitem[Eisenberger et~al.(2020)Eisenberger, Lahner, and Cremers]{eisenberger2020smooth}
Marvin Eisenberger, Zorah Lahner, and Daniel Cremers.
\newblock Smooth shells: Multi-scale shape registration with functional maps.
\newblock In \emph{Proceedings of the IEEE/CVF Conference on Computer Vision and Pattern Recognition}, pages 12265--12274, 2020.

\bibitem[Fischer and Ritschel(2022)]{fischer2022metappearance}
Michael Fischer and Tobias Ritschel.
\newblock Metappearance: Meta-learning for visual appearance reproduction.
\newblock \emph{ACM Transactions on Graphics (TOG)}, 41\penalty0 (6):\penalty0 1--13, 2022.

\bibitem[Fischer and Ritschel(2023)]{fischer2023plateau}
Michael Fischer and Tobias Ritschel.
\newblock Plateau-reduced differentiable path tracing.
\newblock In \emph{Proceedings of the IEEE/CVF Conference on Computer Vision and Pattern Recognition}, pages 4285--4294, 2023.

\bibitem[Fischler and Bolles(1981)]{fischler1981random}
Martin~A Fischler and Robert~C Bolles.
\newblock Random sample consensus: a paradigm for model fitting with applications to image analysis and automated cartography.
\newblock \emph{Communications of the ACM}, 24\penalty0 (6):\penalty0 381--395, 1981.

\bibitem[Fi{\v{s}}er et~al.(2016)Fi{\v{s}}er, Jamri{\v{s}}ka, Luk{\'a}{\v{c}}, Shechtman, Asente, Lu, and S{\`y}kora]{fivser2016stylit}
Jakub Fi{\v{s}}er, Ond{\v{r}}ej Jamri{\v{s}}ka, Michal Luk{\'a}{\v{c}}, Eli Shechtman, Paul Asente, Jingwan Lu, and Daniel S{\`y}kora.
\newblock Stylit: illumination-guided example-based stylization of 3d renderings.
\newblock \emph{ACM Transactions on Graphics (TOG)}, 35\penalty0 (4):\penalty0 1--11, 2016.

\bibitem[Gao et~al.(2022)Gao, Gao, He, Lu, Xu, and Li]{gao2022nerf}
Kyle Gao, Yina Gao, Hongjie He, Dening Lu, Linlin Xu, and Jonathan Li.
\newblock Nerf: Neural radiance field in 3d vision, a comprehensive review.
\newblock \emph{arXiv preprint arXiv:2210.00379}, 2022.

\bibitem[Gatys et~al.(2015)Gatys, Ecker, and Bethge]{gatys2015neural}
Leon~A Gatys, Alexander~S Ecker, and Matthias Bethge.
\newblock A neural algorithm of artistic style.
\newblock \emph{arXiv preprint arXiv:1508.06576}, 2015.

\bibitem[Haque et~al.(2023)Haque, Tancik, Efros, Holynski, and Kanazawa]{haque2023instruct}
Ayaan Haque, Matthew Tancik, Alexei~A Efros, Aleksander Holynski, and Angjoo Kanazawa.
\newblock Instruct-nerf2nerf: Editing 3d scenes with instructions.
\newblock \emph{arXiv preprint arXiv:2303.12789}, 2023.

\bibitem[He et~al.(2019)He, Liao, Chen, Yuan, and Sander]{he2019progressive}
Mingming He, Jing Liao, Dongdong Chen, Lu~Yuan, and Pedro~V Sander.
\newblock Progressive color transfer with dense semantic correspondences.
\newblock \emph{ACM Transactions on Graphics (TOG)}, 38\penalty0 (2):\penalty0 1--18, 2019.

\bibitem[Hedlin et~al.(2023)Hedlin, Sharma, Mahajan, Isack, Kar, Tagliasacchi, and Yi]{hedlin2023unsupervised}
Eric Hedlin, Gopal Sharma, Shweta Mahajan, Hossam Isack, Abhishek Kar, Andrea Tagliasacchi, and Kwang~Moo Yi.
\newblock Unsupervised semantic correspondence using stable diffusion.
\newblock \emph{arXiv preprint arXiv:2305.15581}, 2023.

\bibitem[Hertzmann et~al.(2023)Hertzmann, Jacobs, Oliver, Curless, and Salesin]{hertzmann2023image}
Aaron Hertzmann, Charles~E Jacobs, Nuria Oliver, Brian Curless, and David~H Salesin.
\newblock Image analogies.
\newblock In \emph{Seminal Graphics Papers: Pushing the Boundaries, Volume 2}, pages 557--570. 2023.

\bibitem[Huang et~al.(2021)Huang, Tseng, Saini, Singh, and Yang]{huang2021learning}
Hsin-Ping Huang, Hung-Yu Tseng, Saurabh Saini, Maneesh Singh, and Ming-Hsuan Yang.
\newblock Learning to stylize novel views.
\newblock In \emph{Proceedings of the IEEE/CVF International Conference on Computer Vision}, pages 13869--13878, 2021.

\bibitem[Huang et~al.(2022)Huang, He, Yuan, Lai, and Gao]{huang2022stylizednerf}
Yi-Hua Huang, Yue He, Yu-Jie Yuan, Yu-Kun Lai, and Lin Gao.
\newblock Stylizednerf: consistent 3d scene stylization as stylized nerf via 2d-3d mutual learning.
\newblock In \emph{Proceedings of the IEEE/CVF Conference on Computer Vision and Pattern Recognition}, pages 18342--18352, 2022.

\bibitem[Jacq and Herring(2023)]{pytorchstyletransfer}
Alexis Jacq and Winston Herring.
\newblock Neural style transfer, 2023.
\newblock URL \url{https://pytorch.org/tutorials/advanced/neural_style_tutorial.html}.

\bibitem[Jambon et~al.(2023)Jambon, Kerbl, Kopanas, Diolatzis, Drettakis, and Leimk{\"u}hler]{jambon2023nerfshop}
Cl{\'e}ment Jambon, Bernhard Kerbl, Georgios Kopanas, Stavros Diolatzis, George Drettakis, and Thomas Leimk{\"u}hler.
\newblock Nerfshop: Interactive editing of neural radiance fields.
\newblock \emph{Proceedings of the ACM on Computer Graphics and Interactive Techniques}, 6\penalty0 (1), 2023.

\bibitem[Kingma and Ba(2014)]{kingma2014adam}
Diederik~P Kingma and Jimmy Ba.
\newblock Adam: A method for stochastic optimization.
\newblock \emph{arXiv preprint arXiv:1412.6980}, 2014.

\bibitem[Knapitsch et~al.(2017)Knapitsch, Park, Zhou, and Koltun]{knapitsch2017tanks}
Arno Knapitsch, Jaesik Park, Qian-Yi Zhou, and Vladlen Koltun.
\newblock Tanks and temples: Benchmarking large-scale scene reconstruction.
\newblock \emph{ACM Transactions on Graphics (ToG)}, 36\penalty0 (4):\penalty0 1--13, 2017.

\bibitem[Kobayashi et~al.(2022)Kobayashi, Matsumoto, and Sitzmann]{kobayashi2022decomposing}
Sosuke Kobayashi, Eiichi Matsumoto, and Vincent Sitzmann.
\newblock Decomposing nerf for editing via feature field distillation.
\newblock \emph{Advances in Neural Information Processing Systems}, 35:\penalty0 23311--23330, 2022.

\bibitem[Kopanas and Drettakis(2023)]{kopanas2023improving}
Georgios Kopanas and George Drettakis.
\newblock Improving nerf quality by progressive camera placement for unrestricted navigation in complex environments.
\newblock \emph{arXiv preprint arXiv:2309.00014}, 2023.

\bibitem[Kuang et~al.(2023)Kuang, Luan, Bi, Shu, Wetzstein, and Sunkavalli]{kuang2023palettenerf}
Zhengfei Kuang, Fujun Luan, Sai Bi, Zhixin Shu, Gordon Wetzstein, and Kalyan Sunkavalli.
\newblock Palettenerf: Palette-based appearance editing of neural radiance fields.
\newblock In \emph{Proceedings of the IEEE/CVF Conference on Computer Vision and Pattern Recognition}, pages 20691--20700, 2023.

\bibitem[Lee et~al.(2023)Lee, Gupta, Kim, Makwana, Chen, and Feng]{lee2023so}
Keifer Lee, Shubham Gupta, Sunglyoung Kim, Bhargav Makwana, Chao Chen, and Chen Feng.
\newblock So-nerf: Active view planning for nerf using surrogate objectives.
\newblock \emph{arXiv preprint arXiv:2312.03266}, 2023.

\bibitem[Li et~al.(2017)Li, Fang, Yang, Wang, Lu, and Yang]{li2017universal}
Yijun Li, Chen Fang, Jimei Yang, Zhaowen Wang, Xin Lu, and Ming-Hsuan Yang.
\newblock Universal style transfer via feature transforms.
\newblock \emph{Advances in neural information processing systems}, 30, 2017.

\bibitem[Liamheng(2020)]{wctpytorch}
Liamheng.
\newblock Pytorch1.4-wct-universal style transfer, 2020.
\newblock URL \url{https://github.com/liamheng/Pytorch1.4-WCT}.

\bibitem[Liao et~al.(2017)Liao, Yao, Yuan, Hua, and Kang]{liao2017visual}
Jing Liao, Yuan Yao, Lu~Yuan, Gang Hua, and Sing~Bing Kang.
\newblock Visual attribute transfer through deep image analogy.
\newblock \emph{arXiv preprint arXiv:1705.01088}, 2017.

\bibitem[Liu et~al.(2023{\natexlab{a}})Liu, Fischer, and Ritschel]{liu2023learning}
Chen Liu, Michael Fischer, and Tobias Ritschel.
\newblock Learning to learn and sample brdfs.
\newblock In \emph{Computer Graphics Forum}, volume~42, pages 201--211. Wiley Online Library, 2023{\natexlab{a}}.

\bibitem[Liu et~al.(2022)Liu, Cao, Mao, Zhang, Zhang, Keppo, Shan, Qie, and Shou]{liu2022devrf}
Jia-Wei Liu, Yan-Pei Cao, Weijia Mao, Wenqiao Zhang, David~Junhao Zhang, Jussi Keppo, Ying Shan, Xiaohu Qie, and Mike~Zheng Shou.
\newblock Devrf: Fast deformable voxel radiance fields for dynamic scenes.
\newblock \emph{Advances in Neural Information Processing Systems}, 35:\penalty0 36762--36775, 2022.

\bibitem[Liu et~al.(2023{\natexlab{b}})Liu, Zhan, Chen, Zhang, Yu, El~Saddik, Lu, and Xing]{liu2023stylerf}
Kunhao Liu, Fangneng Zhan, Yiwen Chen, Jiahui Zhang, Yingchen Yu, Abdulmotaleb El~Saddik, Shijian Lu, and Eric~P Xing.
\newblock Stylerf: Zero-shot 3d style transfer of neural radiance fields.
\newblock In \emph{Proceedings of the IEEE/CVF Conference on Computer Vision and Pattern Recognition}, pages 8338--8348, 2023{\natexlab{b}}.

\bibitem[Louis(2021)]{deepimageanalogiespytorch}
Ben Louis.
\newblock Deep image analogies pytorch, 2021.
\newblock URL \url{https://github.com/Ben-Louis/Deep-Image-Analogy-PyTorch}.

\bibitem[Lowe(2004)]{lowe2004distinctive}
David~G Lowe.
\newblock Distinctive image features from scale-invariant keypoints.
\newblock \emph{International journal of computer vision}, 60:\penalty0 91--110, 2004.

\bibitem[Luo et~al.(2023)Luo, Dunlap, Park, Holynski, and Darrell]{luo2023diffusion}
Grace Luo, Lisa Dunlap, Dong~Huk Park, Aleksander Holynski, and Trevor Darrell.
\newblock Diffusion hyperfeatures: Searching through time and space for semantic correspondence.
\newblock \emph{arXiv preprint arXiv:2305.14334}, 2023.

\bibitem[Marr and Hildreth(1980)]{marr1980theory}
David Marr and Ellen Hildreth.
\newblock Theory of edge detection.
\newblock \emph{Proceedings of the Royal Society of London. Series B. Biological Sciences}, 207\penalty0 (1167):\penalty0 187--217, 1980.

\bibitem[Mildenhall et~al.(2021)Mildenhall, Srinivasan, Tancik, Barron, Ramamoorthi, and Ng]{mildenhall2021nerf}
Ben Mildenhall, Pratul~P Srinivasan, Matthew Tancik, Jonathan~T Barron, Ravi Ramamoorthi, and Ren Ng.
\newblock Nerf: Representing scenes as neural radiance fields for view synthesis.
\newblock \emph{Communications of the ACM}, 65\penalty0 (1):\penalty0 99--106, 2021.

\bibitem[Morreale et~al.(2021)Morreale, Aigerman, Kim, and Mitra]{morreale2021neural}
Luca Morreale, Noam Aigerman, Vladimir~G Kim, and Niloy~J Mitra.
\newblock Neural surface maps.
\newblock In \emph{Proceedings of the IEEE/CVF Conference on Computer Vision and Pattern Recognition}, pages 4639--4648, 2021.

\bibitem[Morreale et~al.(2023)Morreale, Aigerman, Kim, and Mitra]{morreale2023neural}
Luca Morreale, Noam Aigerman, Vladimir~G Kim, and Niloy~J Mitra.
\newblock Neural semantic surface maps.
\newblock \emph{arXiv preprint arXiv:2309.04836}, 2023.

\bibitem[M{\"u}ller et~al.(2022)M{\"u}ller, Evans, Schied, and Keller]{muller2022instant}
Thomas M{\"u}ller, Alex Evans, Christoph Schied, and Alexander Keller.
\newblock Instant neural graphics primitives with a multiresolution hash encoding.
\newblock \emph{ACM Transactions on Graphics (ToG)}, 41\penalty0 (4):\penalty0 1--15, 2022.

\bibitem[Ng and Henikoff(2003)]{ng2003sift}
Pauline~C Ng and Steven Henikoff.
\newblock Sift: Predicting amino acid changes that affect protein function.
\newblock \emph{Nucleic acids research}, 31\penalty0 (13):\penalty0 3812--3814, 2003.

\bibitem[Nguyen-Phuoc et~al.(2022)Nguyen-Phuoc, Liu, and Xiao]{nguyen2022snerf}
Thu Nguyen-Phuoc, Feng Liu, and Lei Xiao.
\newblock Snerf: stylized neural implicit representations for 3d scenes.
\newblock \emph{arXiv preprint arXiv:2207.02363}, 2022.

\bibitem[Pang et~al.(2023)Pang, Hua, and Yeung]{pang2023locally}
Hong-Wing Pang, Binh-Son Hua, and Sai-Kit Yeung.
\newblock Locally stylized neural radiance fields.
\newblock In \emph{Proceedings of the IEEE/CVF International Conference on Computer Vision}, pages 307--316, 2023.

\bibitem[Rematas et~al.(2014)Rematas, Ritschel, Fritz, and Tuytelaars]{rematas2014image}
Konstantinos Rematas, Tobias Ritschel, Mario Fritz, and Tinne Tuytelaars.
\newblock Image-based synthesis and re-synthesis of viewpoints guided by 3d models.
\newblock In \emph{Proceedings of the IEEE Conference on Computer Vision and Pattern Recognition}, pages 3898--3905, 2014.

\bibitem[Riba et~al.(2020)Riba, Mishkin, Ponsa, Rublee, and Bradski]{eriba2019kornia}
E.~Riba, D.~Mishkin, D.~Ponsa, E.~Rublee, and G.~Bradski.
\newblock Kornia: an open source differentiable computer vision library for pytorch.
\newblock In \emph{Winter Conference on Applications of Computer Vision}, 2020.

\bibitem[Rublee et~al.(2011)Rublee, Rabaud, Konolige, and Bradski]{rublee2011orb}
Ethan Rublee, Vincent Rabaud, Kurt Konolige, and Gary Bradski.
\newblock Orb: An efficient alternative to sift or surf.
\newblock In \emph{2011 International conference on computer vision}, pages 2564--2571. Ieee, 2011.

\bibitem[Schmidt et~al.(2023)Schmidt, Pieper, and Kobbelt]{schmidt2023surface}
Patrick Schmidt, D{\"o}rte Pieper, and Leif Kobbelt.
\newblock Surface maps via adaptive triangulations.
\newblock In \emph{Computer Graphics Forum}, volume~42. Wiley Online Library, 2023.

\bibitem[Sharma et~al.(2023)Sharma, Philip, Gharbi, Freeman, Durand, and Deschaintre]{sharma2023materialistic}
Prafull Sharma, Julien Philip, Micha{\"e}l Gharbi, Bill Freeman, Fredo Durand, and Valentin Deschaintre.
\newblock Materialistic: Selecting similar materials in images.
\newblock \emph{ACM Transactions on Graphics (TOG)}, 42\penalty0 (4):\penalty0 1--14, 2023.

\bibitem[Shi et~al.(1994)]{shi1994good}
Jianbo Shi et~al.
\newblock Good features to track.
\newblock In \emph{1994 Proceedings of IEEE conference on computer vision and pattern recognition}, pages 593--600. IEEE, 1994.

\bibitem[Song et~al.(2023)Song, Choi, Do, Lee, and Kim]{song2023blending}
Hyeonseop Song, Seokhun Choi, Hoseok Do, Chul Lee, and Taehyeong Kim.
\newblock Blending-nerf: Text-driven localized editing in neural radiance fields.
\newblock In \emph{Proceedings of the IEEE/CVF International Conference on Computer Vision}, pages 14383--14393, 2023.

\bibitem[{\v{S}}ubrtov{\'a} et~al.(2023){\v{S}}ubrtov{\'a}, Luk{\'a}{\v{c}}, {\v{C}}ech, Futschik, Shechtman, and S{\`y}kora]{vsubrtova2023diffusion}
Ad{\'e}la {\v{S}}ubrtov{\'a}, Michal Luk{\'a}{\v{c}}, Jan {\v{C}}ech, David Futschik, Eli Shechtman, and Daniel S{\`y}kora.
\newblock Diffusion image analogies.
\newblock In \emph{ACM SIGGRAPH 2023 Conference Proceedings}, pages 1--10, 2023.

\bibitem[Tancik et~al.(2023)Tancik, Weber, Ng, Li, Yi, Wang, Kristoffersen, Austin, Salahi, Ahuja, et~al.]{tancik2023nerfstudio}
Matthew Tancik, Ethan Weber, Evonne Ng, Ruilong Li, Brent Yi, Terrance Wang, Alexander Kristoffersen, Jake Austin, Kamyar Salahi, Abhik Ahuja, et~al.
\newblock Nerfstudio: A modular framework for neural radiance field development.
\newblock In \emph{ACM SIGGRAPH 2023 Conference Proceedings}, pages 1--12, 2023.

\bibitem[Tang et~al.(2023)Tang, Jia, Wang, Phoo, and Hariharan]{tang2023emergent}
Luming Tang, Menglin Jia, Qianqian Wang, Cheng~Perng Phoo, and Bharath Hariharan.
\newblock Emergent correspondence from image diffusion.
\newblock \emph{arXiv preprint arXiv:2306.03881}, 2023.

\bibitem[Tewari et~al.(2022)Tewari, Thies, Mildenhall, Srinivasan, Tretschk, Yifan, Lassner, Sitzmann, Martin-Brualla, Lombardi, et~al.]{tewari2022advances}
Ayush Tewari, Justus Thies, Ben Mildenhall, Pratul Srinivasan, Edgar Tretschk, Wang Yifan, Christoph Lassner, Vincent Sitzmann, Ricardo Martin-Brualla, Stephen Lombardi, et~al.
\newblock Advances in neural rendering.
\newblock In \emph{Computer Graphics Forum}, volume~41, pages 703--735. Wiley Online Library, 2022.

\bibitem[Tumanyan et~al.(2022)Tumanyan, Bar-Tal, Bagon, and Dekel]{tumanyan2022splicing}
Narek Tumanyan, Omer Bar-Tal, Shai Bagon, and Tali Dekel.
\newblock Splicing vit features for semantic appearance transfer.
\newblock In \emph{Proceedings of the IEEE/CVF Conference on Computer Vision and Pattern Recognition}, pages 10748--10757, 2022.

\bibitem[Wang et~al.(2023{\natexlab{a}})Wang, Dutt, and Mitra]{wang2023proteusnerf}
Binglun Wang, Niladri~Shekhar Dutt, and Niloy~J Mitra.
\newblock Proteusnerf: Fast lightweight nerf editing using 3d-aware image context.
\newblock \emph{arXiv preprint arXiv:2310.09965}, 2023{\natexlab{a}}.

\bibitem[Wang et~al.(2022)Wang, Chai, He, Chen, and Liao]{wang2022clip}
Can Wang, Menglei Chai, Mingming He, Dongdong Chen, and Jing Liao.
\newblock Clip-nerf: Text-and-image driven manipulation of neural radiance fields.
\newblock In \emph{Proceedings of the IEEE/CVF Conference on Computer Vision and Pattern Recognition}, pages 3835--3844, 2022.

\bibitem[Wang et~al.(2023{\natexlab{b}})Wang, Jiang, Chai, He, Chen, and Liao]{wang2023nerf}
Can Wang, Ruixiang Jiang, Menglei Chai, Mingming He, Dongdong Chen, and Jing Liao.
\newblock Nerf-art: Text-driven neural radiance fields stylization.
\newblock \emph{IEEE Transactions on Visualization and Computer Graphics}, 2023{\natexlab{b}}.

\bibitem[Wang et~al.(2023{\natexlab{c}})Wang, Han, Habermann, Daniilidis, Theobalt, and Liu]{wang2023neus2}
Yiming Wang, Qin Han, Marc Habermann, Kostas Daniilidis, Christian Theobalt, and Lingjie Liu.
\newblock Neus2: Fast learning of neural implicit surfaces for multi-view reconstruction.
\newblock In \emph{Proceedings of the IEEE/CVF International Conference on Computer Vision}, pages 3295--3306, 2023{\natexlab{c}}.

\bibitem[Wu et~al.(2023)Wu, Sun, Lai, and Gao]{wu2023nerf}
Tong Wu, Jia-Mu Sun, Yu-Kun Lai, and Lin Gao.
\newblock De-nerf: Decoupled neural radiance fields for view-consistent appearance editing and high-frequency environmental relighting.
\newblock In \emph{ACM SIGGRAPH 2023 Conference Proceedings}, pages 1--11, 2023.

\bibitem[Xu and Harada(2022)]{xu2022deforming}
Tianhan Xu and Tatsuya Harada.
\newblock Deforming radiance fields with cages.
\newblock In \emph{European Conference on Computer Vision}, pages 159--175. Springer, 2022.

\bibitem[Yuan et~al.(2022)Yuan, Sun, Lai, Ma, Jia, and Gao]{yuan2022nerf}
Yu-Jie Yuan, Yang-Tian Sun, Yu-Kun Lai, Yuewen Ma, Rongfei Jia, and Lin Gao.
\newblock Nerf-editing: geometry editing of neural radiance fields.
\newblock In \emph{Proceedings of the IEEE/CVF Conference on Computer Vision and Pattern Recognition}, pages 18353--18364, 2022.

\bibitem[Zhang et~al.(2023{\natexlab{a}})Zhang, Herrmann, Hur, Cabrera, Jampani, Sun, and Yang]{zhang2023tale}
Junyi Zhang, Charles Herrmann, Junhwa Hur, Luisa~Polania Cabrera, Varun Jampani, Deqing Sun, and Ming-Hsuan Yang.
\newblock A tale of two features: Stable diffusion complements dino for zero-shot semantic correspondence.
\newblock \emph{arXiv preprint arXiv:2305.15347}, 2023{\natexlab{a}}.

\bibitem[Zhang et~al.(2022)Zhang, Kolkin, Bi, Luan, Xu, Shechtman, and Snavely]{zhang2022arf}
Kai Zhang, Nick Kolkin, Sai Bi, Fujun Luan, Zexiang Xu, Eli Shechtman, and Noah Snavely.
\newblock Arf: Artistic radiance fields.
\newblock In \emph{European Conference on Computer Vision}, pages 717--733. Springer, 2022.

\bibitem[Zhang et~al.(2023{\natexlab{b}})Zhang, Peng, ShenTu, Shuai, Chen, Yu, Bao, and Zhou]{zhang2023dyn}
Shangzhan Zhang, Sida Peng, Yinji ShenTu, Qing Shuai, Tianrun Chen, Kaicheng Yu, Hujun Bao, and Xiaowei Zhou.
\newblock Dyn-e: Local appearance editing of dynamic neural radiance fields.
\newblock \emph{arXiv preprint arXiv:2307.12909}, 2023{\natexlab{b}}.

\end{thebibliography}
}

\clearpage


\twocolumn[
\begin{@twocolumnfalse}
    \centering 
    \Large\textbf{NeRF Analogies: Example-Based Visual Attribute Transfer for NeRFs \\ \vspace{0.1cm} -- Supplemental Materials --} \\ \vspace{0.25cm}
    \normalsize 
    \parbox{\textwidth}{
    \centering
    Michael Fischer\textsuperscript{1}, Zhengqin Li\textsuperscript{2}, Thu Nguyen-Phuoc\textsuperscript{2}, Alja\v{z} Bo\v{z}i\v{c}\textsuperscript{2}, Zhao Dong\textsuperscript{2}, \\ \vspace{0.1cm}
    Carl Marshall\textsuperscript{2}, Tobias Ritschel\textsuperscript{1} \\ \vspace{0.1cm}
    \textsuperscript{1}University College London, \textsuperscript{2}Meta Reality Labs 
    } \vspace{0.2cm}
\end{@twocolumnfalse}
]

\setcounter{section}{0}    
\setcounter{figure}{0}     
\setcounter{table}{0}      

In this supplemental, we will detail additional details on training, ViT setup and experiment protocol that could not be included in the main paper for reasons of brevity. 
We encourage the reader to also view the electronic supplemental where we show animated versions of our method and the baselines.  

\mysection{Implementation Details}{ImplementationDetails}

\mysubsection{Training}{Training}

We use the standard \ac{NeRF} architecture presented in \cite{mildenhall2021nerf}: a fully-connected MLP with 8 layers a 256 neurons, followed by a single layer of 128 neurons and an output layer activated by a Sigmoid function. 
We use the Adam optimizer \cite{kingma2014adam} with a learning rate of \expnum{1}{-4} and a batchsize of 512. 
We found that some of the correspondences that DiNO produces are noisy, i.e., two points on the target mesh might map to two different points in the source \ac{NeRF}. 
We alleviate this by training with the L1 loss, which encourages sparsity. Our total loss thus is a weighted combination of the color loss $\mathcal{L}_c$ (cf. the main text) and the DoG loss \edgeloss
\begin{equation*}
    \mathcal{L} = \mathcal{L}_c + \lambda \, \mathcal{L}_G,
\end{equation*}
where we set $\lambda$ to be zero for the first 20,000 training iterations , and then gradually fade in the edge-loss by increasing $\lambda$ up to 50. 
We train for a total of 60,000 iterations and are able to create a \ac{NeRF} analogy, including the feature extraction process, in less than two hours on a single GPU.

\mysubsection{ViT Setup}{ViTSetup}
We use DiNO-ViT \cite{caron2021emerging} with the \ac{ViT}-8B backbone, with a standard patch size of 8, a stride of 4 pixels and increased resolution, leading to overlapping patches and smoother feature maps. 
For our application, we found it important to be able to produce dense correspondences at pixel granularity, which is why we abstain from using DiNO-v2, as it uses a larger patch size and hence coarser feature granularity. 
To further increase the spatial resolution of the feature maps, we query DiNO on vertically and horizontally translated versions of the image (four subsequent translations by one pixel in -x and -y direction, respectively).
For images of size 400p, this leads to per-image feature maps of resolution 392, with 384 features per pixel.
We also experimented with diffusion (hyper-) features \cite{luo2023diffusion} and tried replacing, fusing and concatenating them to our DiNO-setup. 
This did not significantly improve the correspondence quality, but doubled the required computations (both during feature extraction and cosine-similarity computation), which is why we decided to stick with our high-resolution DiNO features. 
Research on \ac{ViT} features has shown the positional bias to decrease with layer depth, while the semantic information increases \cite{amir2021deep}. 
As we do not necessarily expect semantically related regions to occupy similar image positions, we thus use the output of the deepest (11th) attention layer, specifically, the key-component of the attention maps, which has been shown to correlate well with semantic similarity \cite{sharma2023materialistic, amir2021deep}.

\mysubsection{Evaluation Details}{EvaluationDetails}
For the real-world scenes, we use NeRFStudio \cite{tancik2023nerfstudio} and train their Instant-NGP model on the scenes provided in the main text. 
For all 2D methods that are lifted to 3D, we train an Instant-NGP \cite{muller2022instant} network with standard hyperparameters for 10,000 iterations, by which point convergence has long been achieved. 
Our setup for all metrics and methods is 200 images, sampled randomly from a sphere around the object, and split into 100 images for training, 20 for validation and 80 for testing.
We evaluate on unseen test-views. 
For the CLIP direction consistency metric we rendered 80 images in a circular trajectory around the object, with a constant elevation of $30^{\circ}$. 
The metrics in Tab. 1 are averaged across the set of the seven synthetic object pairs shown in Fig. 6, which were also presented to the participants of the user study. 
We show NeRF analogies on additional object pairs in the electronic supplemental. 

\mysection{Additional Experiments}{experiments}

In addition to the experiments in the main manuscript, we here investigate a range of other design decisions. 
Firstly, we try replacing the compute-heavy DiNO-descriptors by more lightweight SIFT features, computed densely across the image with Kornia \cite{eriba2019kornia}. 
We re-run our birdhouse test-case with SIFT- instead of DiNO-descriptors and find that they do not perform well, presumably due to SIFT not capturing semantic similarities. 
\myfigure{dino_vs_sift_descr}{Comparison between DiNO- and SIFT-features.}

Moreover, we note that our method can work on any input or output modality that can represent color in 3D.
We thus repeat our experiments with \acp{SDF} and transfer the appearance between two \acp{SDF} fitted with NeuS2 \cite{wang2023neus2}. 

\myfigure{sdf}{A semantic transfer between a bowl of apples and a set of tennis balls, both encoded as \acp{SDF}.}

Additionally, we create a \ac{NeRF} analogy on semantically unrelated, but similarly shaped objects. 
We transfer the appearance of an avocado onto an armchair of similar form and see that, while not working perfectly, our method produces a plausible outcome. 

\myfigure{avocado}{Transfer between semantically unrelated objects.}

\end{document}